\documentclass[letterpaper, 10 pt, conference]{ieeeconf}  

\IEEEoverridecommandlockouts                              

\usepackage{cite}
\usepackage{amsmath,amssymb,amsfonts}
\usepackage{graphicx}
\usepackage{textcomp}
\usepackage{xcolor}
\usepackage{orcidlink}
\usepackage{times}
\usepackage{epsfig}
\usepackage{amssymb}
\usepackage{flexisym}
\usepackage{booktabs}
\usepackage{multirow}
\usepackage{array}
\usepackage{tabularx}
\usepackage{ragged2e}
\usepackage{multicol}
\usepackage{subfigure}
\usepackage{gensymb}
\usepackage{tabstackengine}
\usepackage{adjustbox}
\usepackage{etoolbox}
\usepackage{bm}
\usepackage[all]{nowidow}
\usepackage[ruled,vlined]{algorithm2e}

\title{\LARGE \bf
WorldPlanner: Monte Carlo Tree Search and MPC with Action-Conditioned Visual World Models
}

\author{R. Khorrambakht$^{*1}$, Joaquim Ortiz-Haro$^{*1}$, Joseph Amigo$^{1}$, Omar Mostafa$^{1}$, Daniel Dugas$^{2}$,\\ Franziska Meier$^{2}$, and Ludovic Righetti$^{1}$
\thanks{$^{1}$Center for Robotics and Embodied Intelligence, Tandon School of Engineering, New York University, Brooklyn, NY}
\thanks{$^{2}$ FAIR at Meta. Franziska Meier and Daniel Dugas contributed in an advisory capacity.}
}

\DeclareMathAlphabet\mathbfcal{OMS}{cmsy}{b}{n}
\mathcode`*=\string"8000
\begingroup
\catcode`*=\active
\xdef*{\noexpand\textup{\string*}}
\endgroup

\begin{document}

\maketitle

\begin{abstract}
Robots must understand their environment from raw sensory inputs and reason about the consequences of their actions in it to solve complex tasks. Behavior Cloning (BC) leverages task-specific human demonstrations to learn this knowledge as end-to-end policies. However, these policies are difficult to transfer to new tasks, and generating training data is challenging because it requires careful demonstrations and frequent environment resets. In contrast to such policy-based view, in this paper we take a model-based approach where we collect a few hours of unstructured easy-to-collect play data to learn an action-conditioned visual world model, a diffusion-based action sampler, and optionally a reward model. The world model -- in combination with the action sampler and a reward model -- is then used to optimize long sequences of actions with a Monte Carlo Tree Search (MCTS) planner. The resulting plans are executed on the robot via a zeroth-order Model Predictive Controller (MPC). We show that the action sampler mitigates hallucinations of the world model during planning and validate our approach on 3 real-world robotic tasks with varying levels of planning and modeling complexity. Our experiments support the hypothesis that planning leads to a significant improvement over BC baselines on a standard manipulation test environment. 

\end{abstract}

\IEEEpeerreviewmaketitle
\section{Introduction}
\label{sec:intro}

For a robot to operate in real-world environments, it must reason about physical interactions based on raw sensory observations and the actions it takes. A common approach in the literature, known as Behavior Cloning (BC), implicitly encodes this knowledge into a policy by training on offline human teleoperation data \cite{chi2023diffusion,zhao2023learning}. However, such episodic, task-oriented data must meet certain standards (e.g., low entropy \cite{zhu2025should} and sufficient coverage) and requires frequent environment resets, making it prohibitively expensive to collect at scale. In this paper, we adopt a different formulation: we first learn a dynamics and reward model from unstructured play data and then use it to adapt state-of-the-art planning and control algorithms to synthesize trajectories for new tasks.

Recently, with the advent of high-fidelity video generative models \cite{zhu2024sora} and diffusion architectures equipped with efficient sampling strategies \cite{karras2022elucidating}, explicit modeling of environment dynamics directly in raw image space has seen major breakthroughs \cite{zhu2024sora}. In the context of robotic control and planning, action-conditioned variants of these architectures have been proposed \cite{agarwal2025cosmos}, demonstrating long-horizon future state prediction with minimal degradation. Building on this progress, we propose a framework that, instead of memorizing optimal actions as in BC, synthesizes them through a search process in the action/observation distribution of an unstructured play dataset. Importantly, since such data imposes no constraints on entropy or environment resetting, it can be collected far more easily, either via direct teleoperation or automated exploration policies.

In this paper, we show that this play distribution contains action primitives useful for solving new goal-specific tasks. Specifically, we formulate a Monte Carlo Tree Search (MCTS) planner and a zeroth-order Model Predictive Controller (MPC) built around a small auto-regressive diffusion-based visual world model \cite{alonso2024diffusion} and several image-conditioned reward functions. Our model is trained solely on a small play dataset (few hours of play) on a single local GPU within a few days. A compact world model not pretrained on prior data allows us to properly evaluate the feasibility of generating trajectories solely from prior-play data. However, note that our planning formulation is directly applicable to bigger foundation world models as well. We validate our approach on a range of representative real-world manipulation tasks of varying difficulty, involving both rigid and deformable objects.
Our contributions are as follows:
\begin{itemize}
\item We propose a control and planning framework based on learned world models from \emph{unstructured play data}, avoiding the costly requirements of collecting goal-directed optimal demonstrations and enabling the synthesis of behaviors not directly seen in the play data.
\item We propose to discretize the MCTS search space using a stochastic diffusion model capturing the play distribution. This minimizes out-of-distribution rollouts by constraining search in the training distribution of the world model, thus enabling reliable planning.
\item We conduct our study using real hardware/data and demonstrate quantitative/qualitative validations in manipulation tasks with varying levels of complexity.
\end{itemize}


\section{Related Work}
\label{sec:relatedworks}
\subsection{Short Horizon Planning with World Models}
To our knowledge, most contributions focusing on the run-time generation of actions with world models are limited to short-horizon planning, either due to error accumulation in traditional world models with highly compressed latent spaces \cite{Hansen2022tdmpc, georgiev2024pwm, hafner2023dreamerv3} or due to the prohibitive compute requirements of more recent foundation world models \cite{agarwal2025cosmos}. A recent representative example is \cite{assran2025v}, which models the dynamics in the latent space of a foundation video encoder and uses it to formulate a one-step MPC planner. The authors of \cite{assran2025v} also report one-step MPC with Cosmos foundation world model \cite{agarwal2025cosmos}, in which case each control update takes up to several minutes, demonstrating the need for more efficient models in planning. In this paper, we show that a relatively small diffusion-based world model originally proposed for simple Atari games \cite{schrittwieser2020mastering} can instead be used to learn high-fidelity world models for real-world robotic tasks, avoiding the prohibitive computation cost due to its size and ensuring long-horizon simulation due to the diffusion retraction to the image manifold at each prediction step. Additionally, we train our world model from scratch from a small-scale (few hours) dataset, showing the possibility of adoption as a local visual dynamical model for planning and control. 

\subsection{Long Horizon Planning}
The prior research on real-world robotic long-horizon planning with visual world models is highly sparse, with only very recent examples such as \cite{neary2025improvingpretrainedvisionlanguageactionpolicies} that adopt a simulator as the world model and a pretrained vision-language-action model as an action proposal module for the MCTS. Alternative examples include the AlphaGo line of work \cite{silver2016mastering} that proposes the adoption of learning in MCTS in achieving remarkable play performance in game environments. To our knowledge, this paper is the first showing long-horizon multi-step planning in real-world robotic settings using learned world models. It demonstrates that with suitable world and action proposal models, such run-time reasoning and planning solutions may also be adopted to solve robotic tasks. Furthermore, we demonstrate that planning in low-level action space may be rendered feasible by discretizing the action space using a diffusion action model trained on unstructured play datasets. 
\newcommand{\norm}[1]{\left\lVert#1\right\rVert}
\newcommand{\quim}[1]{\textcolor{blue}{#1}}
\section{Method}
\label{sec:method}
Our method consists of two stages. In the first stage, we train a world model, a reward model, and an action prior model from a dataset of unstructured play trajectories. In the second stage, we use these models to implement an MCTS planner and an MPC controller to generate new motions. We present the notation and formal problem definition in Sec.~\ref{sec:notation}, describe the learned models in Secs.~\ref{sec:world-model}, \ref{sec:sampler}, and \ref{sec:reward_models}, and then introduce the MCTS planner in Sec.~\ref{sec:mcts} and the MPC controller in Sec.~\ref{sec:method_local_planner}.

\subsection{Notation and Problem Definition}
\label{sec:notation}
We adopt the standard notation of Markov Decision Processes (MDPs), 
$\mathcal{M} = (\mathcal{S}, \mathcal{A}, \mathcal{P}, r, \rho, \gamma)$, 
where $s \in \mathcal{S}$ denotes the state and $a \in \mathcal{A}$ the action. 
The probability density $\mathcal{P}(s' \mid s,a)$ models the environment dynamics, 
while $\rho(s)$ is the distribution over initial states. 
For each state--action pair, $r(s,a)$ denotes the reward, and the goal is to find a policy $\pi$---in our case, the combination of a global planner and a local controller---that maximizes the discounted return:
\[
    V^\pi(s) = \mathbb{E}_\pi\left[\sum_{t=0}^{H} \gamma^t r(s_t, a_t)\right],
\]
where $s_0 = s$, $a_t \sim \pi(s_t)$, and $s_{t+1} \sim \mathcal{P}(\cdot \mid s_t, a_t)$. 

In this paper, we explicitly learn the environment dynamics $\mathcal{P}(s' \mid s,a)$ from offline data. 
Specifically, we assume access to a dataset of random-play trajectories:
\[
    \mathcal{D} = \{\tau_1, \dots, \tau_M\}, \quad 
    \tau_i = \{(s_0, a_0), \dots, (s_N, a_N)\},
\]
where states $s$ correspond to scene observations (e.g., one or multiple camera views), 
and actions $a$ correspond to commands sent to the robot (desired end-effector velocities in Cartesian space). Importantly, the trajectories in $\mathcal{D}$ are \emph{not} goal-conditioned; rather, they represent a collection of interactions that the robot can perform with the environment.
Our objective is to generate and execute new plans that transition from an initial state $s_s$ to a target state $s_g$.

\subsection{Diffusion World Model}
\label{sec:world-model}
We adopt the formulation proposed by the task-specific DIAMOND Atari world model \cite{alonso2024diffusion} due to its high visual fidelity and its fast training and inference on a local GPU. 
Furthermore, because DIAMOND is a diffusion generative model, it can capture the multi-modality of play behavior.

The goal of this world model is to capture the distribution of future states $s_{t+1}$ conditioned on a history of past actions and states $(s_{t:t-h}, a_{t:t-h})$. Specifically, we train a denoiser $D_\theta$ parametrized by $\theta$ and trained by optimizing the following score-matching loss \cite{alonso2024diffusion}:
\begin{equation}
    \label{eq:score_matching_loss}
    \mathcal{L}(\theta)=\norm{D_\theta(s_{t+1}^\tau|s_{t:t-h}, \ a_{t:t-h})-s_{t+1}}
\end{equation}

where $s^\tau_{t+1}$ is the noisy version of the clean $s_{t+1}$ at diffusion step $\tau$. During inference, we sample the next state by iteratively solving the reverse diffusion process. It is important to note that planning and control require the quick generation of many parallel rollouts; thus, denoiser evaluation in sampling the next states has to be kept as small as possible. The noise scheduling, network preconditioning, and integration paradigm adopted by DIAMOND is based on EDM \cite{karras2022elucidating} and enables high-quality sampling only with a few (3) denoising steps.

The denoiser is realized using a U-Net \cite{ronneberger2015u} conditioned on the image history stacked channel-wise and concatenated with the noisy image input, while action and diffusion step conditioning is realized through adaptive group normalization \cite{zheng2020learning}. If the setup is comprised of more than one camera views, the RGB images are stacked channel-wise and treated as a single image with $n_{view}\times3$ channels.

Note that our formulation is aligned with recent works on learning foundation world models in robotics, e.g., \cite{agarwal2025cosmos, bruce2024genie}. 
In this paper, we opt to use a small task-specific world model to properly investigate our dataset-driven hypothesis: that tasks can be solved using the knowledge embedded in unstructured and goal-independent play datasets, while also reducing computational constraints (training and inference on a desktop GPU). 
Furthermore, a small world model could serve as a local world model distilled from a larger model to accelerate online planning.

\subsection{Action Generator Model}
\label{sec:sampler}
The distribution of robot actions in the unstructured play dataset $\mathcal{D}$ contains meaningful interactions and motion primitives that can later be combined through planning to solve new tasks. 
Thus, in this project, we also learn a play action policy, denoted as $\pi_{prior}(a_t \mid s_t)$, using the same play dataset $\mathcal{D}$. 
Specifically, we use a diffusion policy \cite{chi2023diffusion} to achieve diversity and multimodality. 
Together with the world model, the stochastic action policy enables sampling of short trajectories of action/state pairs $a_{t:{t+H}}, s_{t:{t+H}}$, which will be used in Sec.~\ref{sec:mcts} to expand the search tree. 
An additional benefit of learning the action distribution is that the sampled actions remain within the training distribution of the world model, thereby minimizing the probability of hallucination. 

\subsection{Reward Models}
\label{sec:reward_models}
Evaluating the rollouts generated during planning and control in the world model will require an image-conditioned reward function. In this paper, we consider three approaches to formulate image-space reward functions:
\subsubsection{Fully Geometric} 
\label{sec:reward_geometric}
For tasks where the objective is explicitly representable geometrically, an off-the-shelf object pose tracker (position and orientation) can be adopted as part of the reward function. Specifically, the tracker $f_\psi(s_t)$ is incorporated as input to an explicit reward function $r(f_\psi(s_t), a_t)$ that formulates a geometric objective of the task (e.g. position difference). 

\subsubsection{Latent Space Image Distance with Pretrained Visual Foundation Models}
\label{sec:reward_dino}
A more general approach represents the goal as an image $s_{goal}$ and encodes both the current and goal images into a high-dimensional embedding space, where the geometric distance (e.g., cosine similarity or Euclidean) reflects the semantic or geometric difference between the two images, defined as:
\begin{equation}
    \label{eq:dino_reward}
    r(s_t, a_t| s_{goal}) = r_a(a_t) - \alpha \norm{f_\psi(s_t)-f_\psi(s_{goal})}
\end{equation}
where $r_a(\cdot)$ is an action-dependent reward term (e.g., the squared norm of the action), and $\alpha$ is a tunable hyperparameter. 
In our experiments, $f_\psi(s_t) : \mathbb{R}^{W \times H \times C} \to \mathbb{R}^d$ is a pretrained foundation image encoder (DINOv2 \cite{oquab2023dinov2}). 
We additionally propose adopting an optional object-centric attention mechanism that retains only the object(s) of interest in the image before computing the $f_\psi(\cdot)$ embedding. 
In this case, the DINOv2 distance is ensured to encode only the dissimilarity of the objects of interest between the query and goal images. 
We empirically observed that this filtering improves planning performance and convergence rate.

\subsubsection{Training Reward Functions from Video}
\label{sec:rand2reward}
A reward model encoding temporal progress can be trained on a passive dataset of observations of the robot interacting with the environment during play or while performing specific tasks. 
Specifically, given a start and goal state $s_{start}, \ s_{goal}$ sampled from temporal sequences of states, the objective is to learn an energy function 
$f_\psi(s_t \mid s_{start}, s_{goal})$
that assigns higher values to states $s_t$ that are temporally closer to $s_{goal}$. 
We adopt the Bradley--Terry $\mathcal{L}(\psi)$ objective from \cite{yang2024rank2reward} to train this reward model:
\begin{equation}
\label{eq:rank2reward}
\begin{split}
\mathcal{L}(\psi) &= 
\frac{e^{f_\psi(s_i)}}{e^{f_\psi(s_i)} + e^{f_\psi(s_j)}} \, \mathbf{1}\{ i > j \} \\
&\quad + \frac{e^{f_\psi(s_j)}}{e^{f_\psi(s_i)} + e^{f_\psi(s_j)}} \, \mathbf{1}\{ i \leq j \},
\end{split}
\end{equation}
where $s_i$ and $s_j$ are states randomly sampled from a video chunk whose first and last frames are $s_{start}$ and $s_{goal}$. 
The learned energy function $f_\psi$ takes DINOv2 features as input instead of raw images. 
For clarity, $s_{start}$ and $s_{goal}$ are omitted from the notation of the energy function $f_\psi(\cdot \mid s_{start}, s_{goal})$ in Eq.~\eqref{eq:rank2reward}.

\begin{algorithm}[bt]
\caption{Global Planner: (MCTS)}
\label{alg:mcts}
\DontPrintSemicolon
\KwIn{Initial state $s_0$, prior policy $\pi_{prior}$, world model $\mathcal{P}$, reward model $r$, horizon $H_{sim}$, rollout number $M_{sim}$, exploration constant $c$, minimum visit count for solution node $n_{min}$}
\KwOut{Plan from root to node with maximum average value}
Initialize root node $v_0 \gets s_0$ with $n_{visit}(v_0)=0$, $V_{total}(v_0)=0$\;
Expand $v_0$ using $\pi_{prior}$ and $\mathcal{P}$\;
\While{termination criterion not reached}{
  \tcp{Selection}
  Traverse from $v_0$ to a leaf by choosing child with maximum UCB1 (Eq.~\eqref{eq:ucb1})\;
  
  \tcp{Expansion}
  \If{$v$ visited}{Expand via $\pi_{prior}, \mathcal{P}$; set $v$ to first new child}
  
  \tcp{Simulation}
  From $v$, run $M_{sim}$ rollouts to horizon $H_{sim}$ using $\pi_{prior}$ and $\mathcal{P}$; compute $R$ via Eq.~\eqref{eq:mcts_game_score}\;
  
  \tcp{Backpropagation}
  For each node $u$ on path $v \to v_0$: update $V_{total}(u)\!\gets\!V_{total}(u)+R$, $n_{visit}(u)\!\gets\!n_{visit}(u)+1$\;
}
\Return Plan from root to node with maximum average value and $n_{visit} > n_{min}$.\;
\end{algorithm}

\subsection{Global MCTS Planner}
\label{sec:mcts}
In this section, we formulate the MCTS algorithm using the learned world model as a simulator. 
The goal of the planner is to take the robot from its initial state $s_s$ to a goal state $s_g$. 
Starting from the initial state, MCTS builds a search tree where the nodes of the tree are states, and the edges are short trajectories generated by rolling out the prior policy $\pi_{prior}$.

The tree is built using the classical MCTS steps:
traversal, expansion, simulation, and backpropagation.

\textit{1)} In the \textit{traversal step}, the search begins at the root and identifies the most promising node to evaluate or expand. 
The notion of “best” is quantified through the Upper Confidence Bound (UCB1), calculated as follows for each node:

\begin{equation}
\label{eq:ucb1}
UCB1(node) = \frac{V_{total}}{n_{visit}} + c\sqrt{\frac{\log(N)}{n_{visit}}}
\end{equation}
where $n_{visit}$ is the number of visits to the node, $N$ is the total number of visits to the parent node, $c$ is a constant determining the exploration–exploitation balance, and $V_{total}$ is the estimate of the value function.


\textit{2)} For the \textit{expansion step}, we evaluate the prior action policy starting from a leaf node. 
The policy generates a set of short sequences of actions, which are rolled out in the simulator in parallel. 
The number of parallel rollouts defines the branching factor of the tree and is an important user-defined parameter. 
The last state of each rollout is added to the tree.

\textit{3)}
The \textit{simulation step} starts from one of the newly added nodes. 
We generate $M_{sim}$ rollouts of the prior policy in parallel for a longer, fixed number of steps $H_{sim}$
The simulation score is then:
\begin{equation}
    \label{eq:mcts_game_score}
    \max_{\substack{T \in \{0 \cdots H_{sim}\} \\ k \in \{0 \cdots M_{sim}\}}} \sum_{t}^{T} r(s_t^k, a_t^k).
\end{equation}

Each rollout leverages the stochasticity of both the world model and the diffusion policy to explore the various plausible action modalities for the given state. 

    \textit{4)}
Finally, in the \textit{backpropagation step}, the attained score is used to update the total reward for the nodes on the path leading to the chosen leaf node, by accumulating the total reward for each of those nodes and incrementing their visit counters. 

The four MCTS steps are repeated until a maximum number of iterations is reached or a node state with a desired average value and $n_{visit}$ larger than a user-defined threshold $n_{min}$ is found.  
This procedure is summarized in Alg.~\ref{alg:mcts}.

\begin{algorithm}[bt]
\caption{Local Zeroth-Order MPC Controller}
\label{alg:mpc}
\DontPrintSemicolon
\KwIn{MCTS plan $\tau^{MCTS}$, window $H_{MPC}$, execution chunk $h_{\text{exec}}$, populations $K,K'$, elites $K_{elite}$, iterations $N_{MPC}$, distance metric $d$, max deviation $d_{max}$}
\KwOut{Executed action sequence with visual feedback corrections}
Initialize time index $t\gets 0$\;
\While{$t < H_P$}{
  \tcp{Population Generation}
  Around planned actions in window $\{t,\dots,t+H_{MPC}\}$, define Gaussians $\mathcal{N}(\mu_t,\Sigma_t)$ with $\mu_t=a_t^P$\;
  Sample $K$ trajectories $A^1,\dots,A^K$ and add $K'$ zero-mean trajectories\;
  
  \tcp{Iterations}
  \For{$i \gets 1$ \KwTo $N_{MPC}$}{
      \tcp{Rollout and Evaluation}
      For each $A^k$, rollout with $\pi_{prior}$ and $\mathcal{P}$ from $s_{\text{cam}}$; compute reward $R^k$\;
      
      \tcp{Update}
      Select top-$K_{elite}$ samples by $R^k$\;
      Update $\mu_t, \Sigma_t$ via weighted mean/variance (Eq.~\eqref{eq:mpc-update}), enforce diagonal $\Sigma_t$ and minimum variance\;
      \tcp{Resample}
      Sample $K+K'$ action trajectories using the updated distribution parameters. 
  }
  
  \tcp{Execution}
  Execute first $h_{\text{exec}}$ actions on robot; capture new $s_{\text{cam}}$;
  Update $t$.
  
  \tcp{Termination Check}
  \If{$d(s_{\text{cam}},s_t^P) > d_{max}$}{Trigger MCTS re-planning; \textbf{break}}
}
\end{algorithm}
\subsection{Local MPC Controller}
\label{sec:method_local_planner}
Open-loop execution of the plan from the previous section may deviate from the desired motion due to cumulative noise and model errors. 
Addressing this problem requires closing the visual feedback loop, either by distilling the planner into a neural policy or by taking the plan as input to an image-based MPC formulated using the same world and value models. 
In this paper, we focus on the latter solution as it enables direct online movement generation.

We use a zeroth-order optimizer \cite{jordana2025introduction} to leverage cost smoothing, improved performance in avoiding local minima compared to gradient-based counterparts (especially for neural components with noisy gradients), and to avoid the need for computing gradients from the world and reward models. Further, this approach can easily be parallelized on GPUs.

Starting at $t=0$, and a plan from MCTS (i.e., a sequence of actions and states) $\tau^{MCTS}=\{(a_0^P,s_0^P), \cdots (a_{h_P}^P,s_{h_P}^P)\}$ with length $H_P$ we define the procedure as follows:
\subsubsection{Population Generation} 
\label{sec:population_generation}
Around each planned action in the optimization window $\{t, \cdots , t+H_{MPC}\}$, we define a Gaussian distribution $a^k_t \sim \mathcal{N}(\mu_t, \Sigma_t)$ with mean $\mu_t=a_t \in \mathbb{R}^{n_u} $ and $\Sigma_t=diag(\{\sigma_1, \cdots ,\sigma_{n_u}\}) \in \mathbb{R}^{n_u \times n_u}$ where $n_u$ is the number of control inputs. We then draw $K$ action samples from each distribution to construct $K$ action trajectories $\mathcal{A}^{plan}=\{A^1, \cdots, A^K\}$ where $A^k=\{a_t^k, \cdots a_{t+H_{MPC}}^k\}$ and $a_t^k \sim \mathcal{N}(\mu_t, \Sigma_t)$. Additionally, we augment $\mathcal{A}^{plan}$ with $\mathcal{A}^0$ comprised of $K'$ extra action trajectories with $\mu=0$ (zero mean) and $\Sigma_0$ set to a fixed predefined value to further enrich the initial action population.
\subsubsection{Rollout and Evaluation}
Using the world and value models, along with $s_{\text{cam}}$ captured from the robot cameras, we propagate each action sample from $\mathcal{A}^{plan} \cup \mathcal{A}^0$ and evaluate the cost/value for each rollout to yield $\mathcal{R}=\{R^1, \cdots, R^{K+K'}\}$, where $R^k$ is the reward/cost associated with sample $k$ from the action population and combines an action cost with a plan-following cost.

\subsubsection{Update} Given each $A^k$ from the population and the corresponding reward $R^k$, we can now update the plan along the MPC horizon. While there are various ways of updating the nominal action \cite{jordana2025introduction}, here we adopt the CEM update rule as used in \cite{hansen2022temporal} where we update the distribution parameters based on a subset of $K_{elite}$ action samples corresponding to the top-$K_{elite}$ best rewards. Specifically, we update the $\mu_t$ and $\Sigma_t$ by computing the sample mean and variance for each individual timesteps as follows:
\begin{equation}
    \label{eq:mpc-update}
    \begin{split}
        &\mu_t \leftarrow \frac{\sum_{m \in \ elites} R^m a^m_t}{\sum_{m \in \ elites} R^m}, \\
        &\Sigma_t \leftarrow \frac{\sum_{m \in \ elites} R^m (a^m_t-\mu_t)(a^m_t-\mu_t)^\top}{\sum_{m \in \ elites} R^m}
    \end{split}
\end{equation}
For simplicity, we force the off-diagonal elements of the covariance matrix to zero, and to avoid exploration collapsing to zero, we enforce a minimum diagonal covariance at each step. 
Note that if we set $K_{elite}=1$ and $\Sigma_t$ to a fixed baseline $\Sigma_0$, we recover the predictive sampling method \cite{howell2022predictive}. 
Also note that optimizing directly over actions may produce non-smooth results, so in practice we optimize over the knot points of a smooth curve \cite{howell2022predictive}. 

We repeat the sampling and evaluation steps for a fixed $N_{MPC}$ iterations and execute an action sequence of length $h_{\text{exec}}$ on the robot. 
We then take the new observation $s_{\text{cam}}$, compute the best actions to follow the next part of the plan, run them on the robot, and continue the process until the goal is reached. 
If the execution deviates too much from the plan, we stop the control and call the MCTS planner again.

\begin{figure}[t]
    \centering 
    \includegraphics[width=0.8\linewidth]{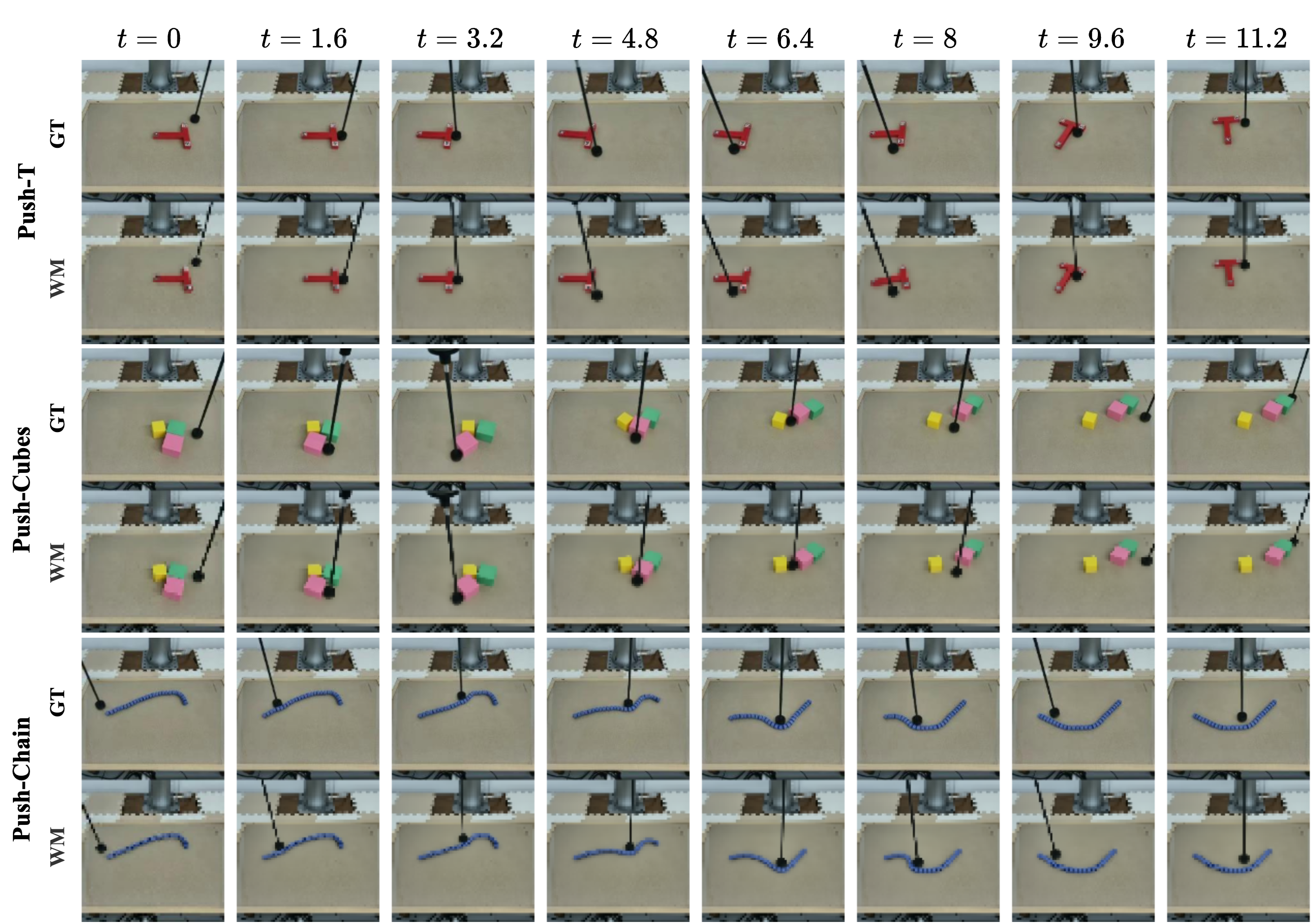}
    \caption{World model rollouts vs ground truth for all the tasks studied.}
    \label{fig:rollouts}
\end{figure}

\section{Results}
\label{sec:results}
\subsection{Setup and Training Details}
All our models are trained on real-world data collected with a 7-DoF robotic arm (Flexiv Rizon-10S). Specifically, the robot is teleoperated by a user to perform unstructured and randomized interactions with the scene for roughly four hours. Note that this dataset is allowed to have high entropy (which is not desirable in BC datasets \cite{zhu2025should}) and does not require environment resets, making it much easier to collect compared to task-specific behavior cloning datasets. Using this data, we train the world model, the prior play policy $\pi_{prior}$, and the reward model described in Sec.~\ref{sec:method}.

The world model is trained on a single NVIDIA RTX 4090 GPU with 24 GB of RAM and requires roughly three days to achieve acceptable long-horizon rollouts. For the action prior, we use the LeRobot \cite{cadene2024lerobot} implementation of the diffusion policy and observe stable random exploration behavior after less than one hour of training. Finally, the reward function $f_\psi(.|s_{start}, s_{goal})$ described in Sec.~\ref{sec:rand2reward} is implemented as a ViT \cite{dosovitskiy2020image} with learned positional embeddings. It takes as input the batch embeddings of the start, goal, and query images, and predicts the corresponding value as output. This encoder is trained by maximizing Eq.~\eqref{eq:rank2reward} on random states drawn from interaction video sequences of length $12.8$ s, with convergence observed after roughly $12$ hours. Additionally, the geometric object tracker used in Sec.~\ref{sec:reward_geometric} is a CNN trained on a small independent robot-object interaction dataset with labels provided from an AprilTag-based pose tracker. 

\begin{figure}[tb]
    \centering
    \includegraphics[width=0.8\linewidth]{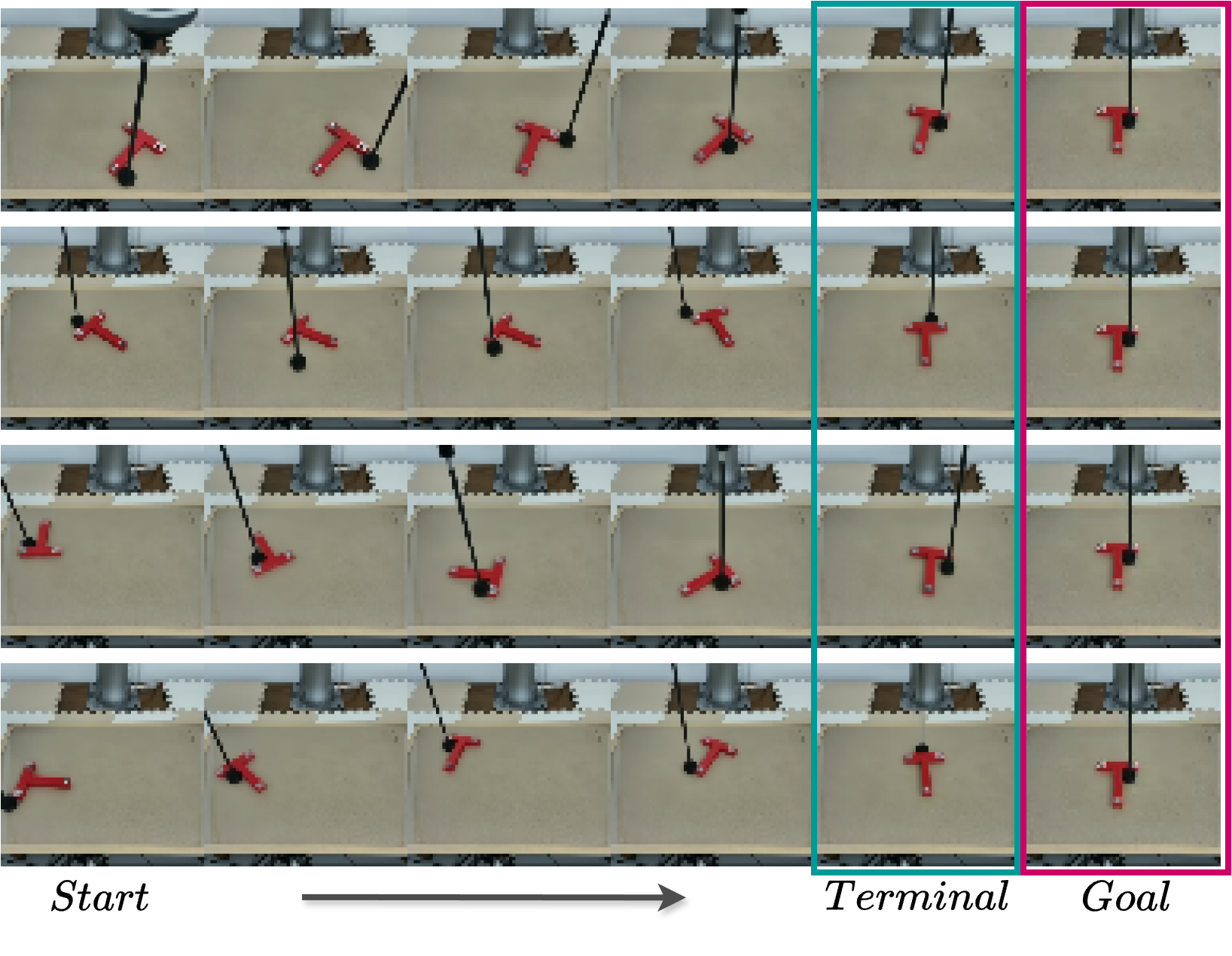}
    \caption{Examples of the MCTS plans for the push-T task aiming to push the object to the center of the board.}
    \label{fig:pusht-planner-rollout}
\end{figure}

\subsection{Environments and Corresponding World Models}

We evaluate our method for three categories of tasks with evolving degrees of modeling and planning complexity. The \emph{push-T} task, being the simplest, requires capturing the interaction between the robot and a single rigid-body object sliding on a surface. The next level is the \emph{push-cubes} tasks that additionally includes inter-object collisions. Finally, the \emph{push-chain} task considers manipulating deformable objects. 
Together, these tasks demonstrate that with the same world model formulation, a wide variety of complex real-world dynamic effects can be modeled. Fig.~\ref{fig:rollouts} shows the trained world model rollouts for each of these environments. The comparison between the Ground Truth (GT) and World Model (WM) rollouts in Fig.~\ref{fig:rollouts} demonstrates that the predictions remain highly consistent with the state of the world even after $11.28$ seconds ($64$ steps) of auto-regressive forward integration. 

\subsection{Single-Object/Robot Interaction}
\label{sec:single_object_exp}
We consider the task of pushing a T-shaped tool from a random initial pose to the center of the board as a simple controlled environment. This task represents the complexities of contact modeling and sequential decision making while being structured and simple enough to evaluate quantitatively using geometrical metrics (Sec.\ref{sec:reward_geometric}). Qualitative examples of the generated plans are demonstrated in Fig.~\ref{fig:pusht-planner-rollout} and show the planner's ability in determining sequences of contact points and motions that solve the task. 

Next, we conduct a quantitative study to compare the MCTS planner against BC baselines on 100 random initializations of the T-tool pose. To ensure identical initialization for all baselines, we use the world model as a simulator and sample 100 random states of the board from the play dataset as start states. We also compare our MCTS planner with two types of rewards against two BC policies based on transformers (ACT \cite{zhao2023learning}) and diffusion modeling (diffusion policy \cite{chi2023diffusion}). The BC baselines are trained on fewer than 100 high-quality human demonstrations independent from the play dataset used to train the world model. Each of the two MCTS baselines respectively employs the geometrical reward in Sec.~\ref{sec:reward_geometric}, and the video reward model in Sec.~\ref{sec:rand2reward}. Notably, the latter is trained on the same demonstration dataset used to train the BC baselines (only passive video observations used to train the value model) to maintain fairness in comparisons against the BC baselines.

For each random starting configuration, we evaluate the BC policies five times for a duration equal to the longest plan generated by MCTS. For both the planner and the BC baselines, a trial is considered successful if, at any point during the plan, there exists a state with a yaw rotation error smaller than $0.3$ radians and translation errors smaller than $e_d \in \{2.5, 5, 7.5, 10\}$ centimeters. The results are shown in Table~\ref{tab:bc_baselines}, where MCTS achieves a higher success rate across all threshold levels. We hypothesize that this improvement is due to the planner’s ability to generate new behaviors online, even if they were not demonstrated in the training data. In contrast, a BC policy simply memorizes demonstrations and, when encountering mistakes, cannot inherently recover if corrective behaviors were absent during training

\begin{figure}[t]
    \centering
    \includegraphics[width=0.85\linewidth]{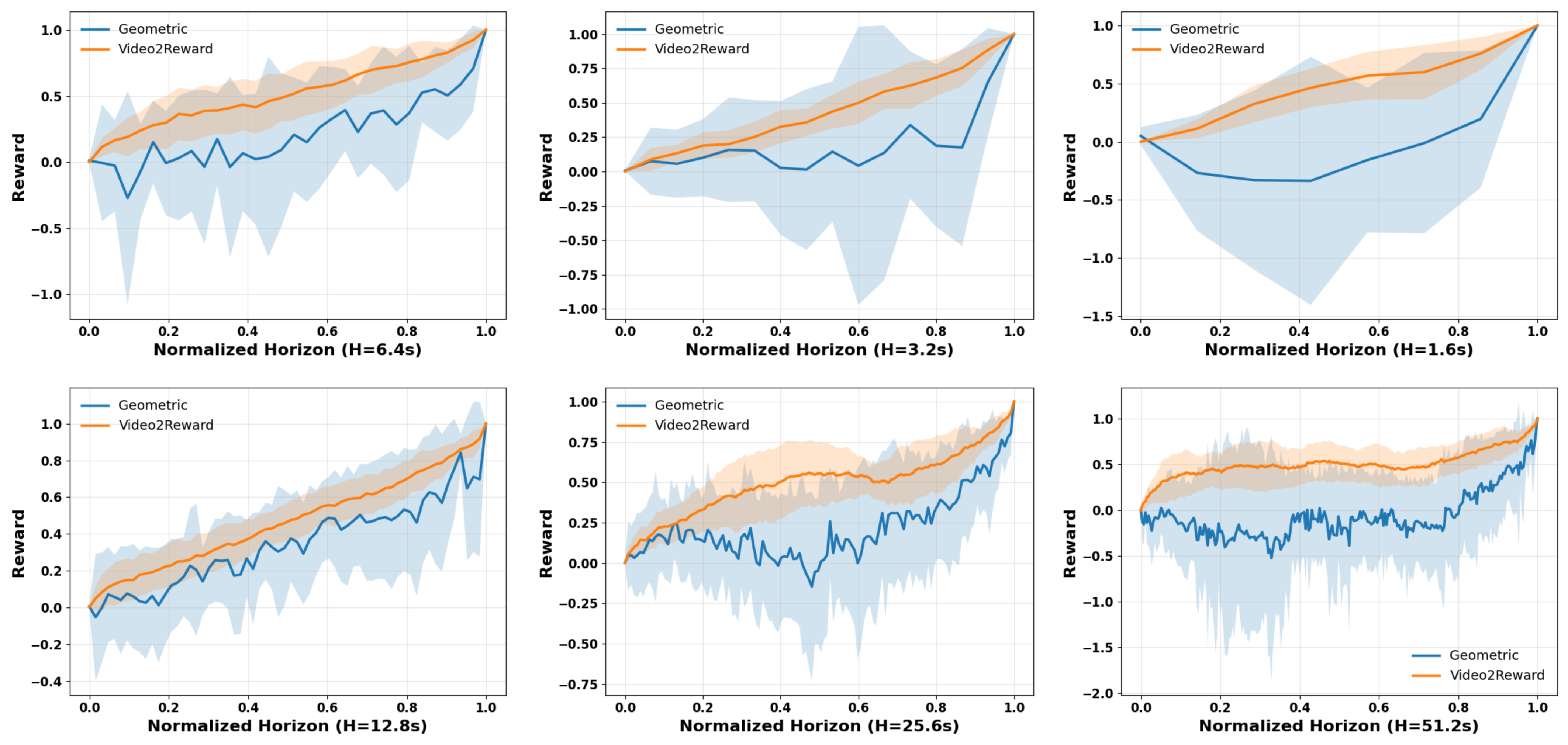}
    \caption{The evolution of the normalized reward predicted by the geometrical reward model (Sec.~\ref{sec:reward_geometric}) and the reward learned from passive videos (Sec.~\ref {sec:rand2reward}) on 10 random selections of 6 temporal lengths.}
    \label{fig:video2reward-progress}
\end{figure}

Running the MCTS using the video to reward model in Sec.~\ref {sec:rand2reward} leads to the highest performance.
Fig.~\ref{fig:video2reward-progress} shows the value distribution predicted by the learned reward model on sequences of different temporal lengths (10 random runs per sequence length). As can be seen, the learned reward model exhibits clear, low-variance monotonic progress as the query state approaches the goal, while the geometric metric only considers the geodesic distance to the goal and does not assign value to the intermediate transition states required to reach the goal state. Note that, as expected, when the trial sequence lengths exceed the video lengths ($12.8$ s) used to train the model, the predictions deviate from a monotonic linear trend but still remain approximately monotonic.

\begin{table}[t]
    \centering
    \caption{Success rates (\%) of MCTS compared to BC baselines on 100 random settings of the push-T task.}
    \label{tab:bc_baselines}
    \resizebox{\columnwidth}{!}{%
    \begin{tabular}{lcccc}
        \toprule
        Method           & $\leq 2.5$\,cm & $\leq 5$\,cm & $\leq 7.5$\,cm & $\leq 10$\,cm \\
        \midrule
        MCTS+Video Reward                                 & 69\% & 92\% & 95\% & 97\% \\
        MCTS+Geometric Reward                             & 52\% & 81\% & 86\% & 91\% \\
        Diffusion Policy$^{*}$ \cite{chi2023diffusion}    & 49\% & 70\% & 83\% & 88\% \\
        ACT$^{*}$ \cite{zhao2023learning}                 & 32\% & 54\% & 57\% & 60\% \\
        \bottomrule \\
    \end{tabular}}
    \\
    $^{*}$Diffusion Policy and ACT are trained on goal-directed data, while the world model used with MCTS is trained on play data. 
\end{table}

\subsection{Multi-Object Selective Interaction}
\begin{figure}[bt]
    \includegraphics[width=0.9\linewidth]{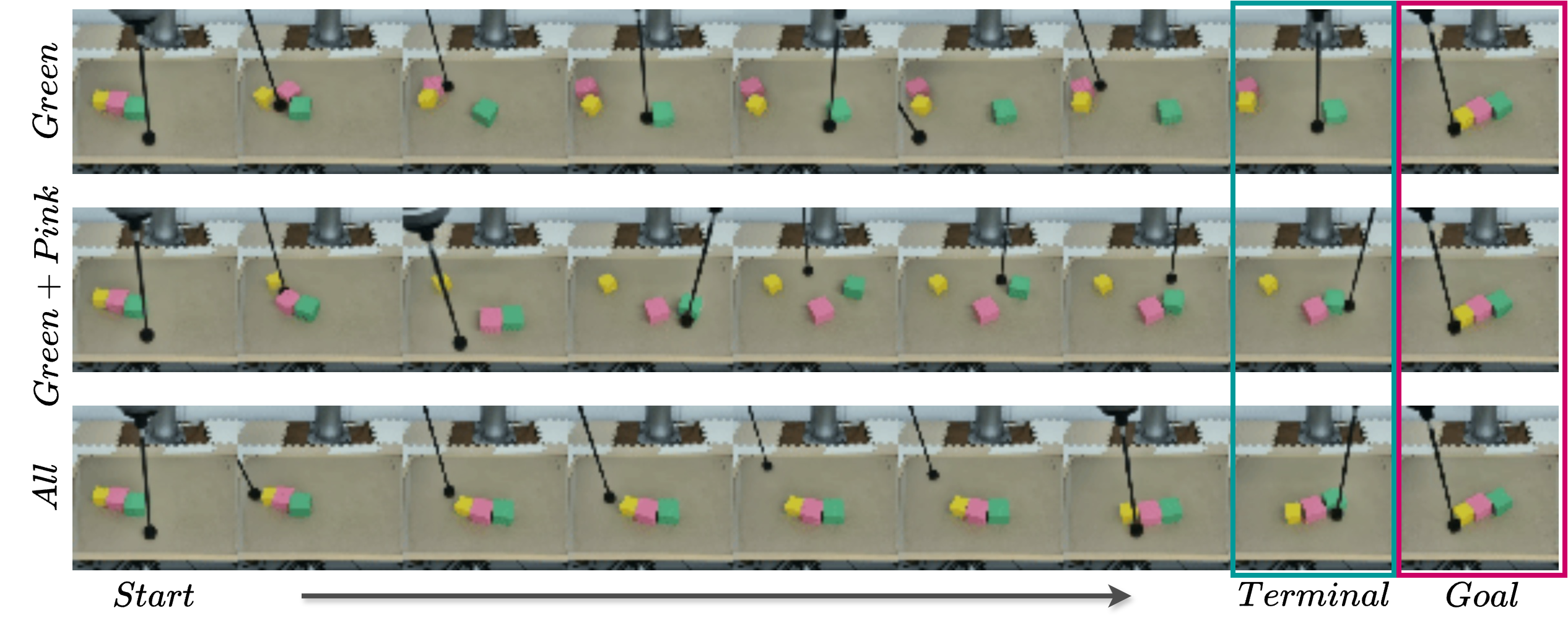}
    \caption{The plans generated with the world model and MCTS. The inter-object contacts are leveraged to move the cubes, while the masked DINOv2 reward model enables generating plans for selective manipulation of objects of interest in the scene. }
    \label{fig:cubes-qualitative}
\end{figure}

One advantage of training a world model and using planning is the ability to optimize for diverse objectives. Here, we show that by changing the reward, we can generate plans to solve different tasks: either moving three cubes to the target position or moving a single cube. To modulate this objective, we use the masked DINOv2-based reward model presented in Sec.~\ref{sec:reward_dino}, where the object of interest is selected in the image using the SAM2 model \cite{ravi2024sam}, and the reward is computed from the embedding distances between the masked states. With this reward, the planner is capable of manipulating the object of interest until it matches its corresponding configuration in the goal image. Importantly, in the generated plans, other objects may also be used if they help indirectly push the object of interest. Fig.~\ref{fig:cubes-qualitative} shows a qualitative visualization of plans for this task.

\begin{figure*}[tb]
    \begin{center}
        \includegraphics[width=0.95\linewidth]{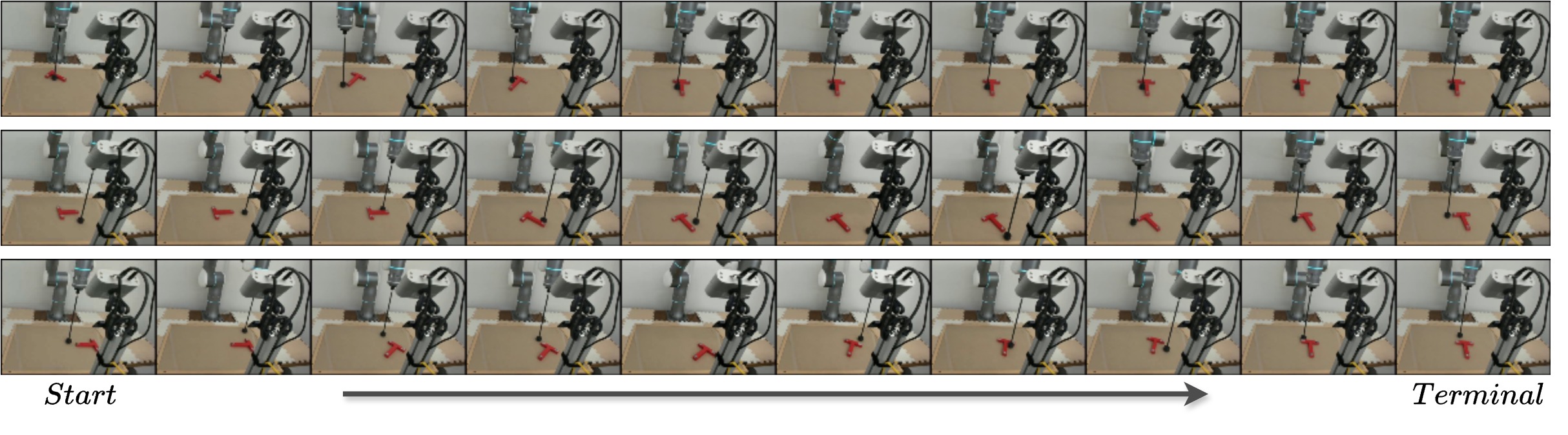}
        \caption{MCTS plans executed on the real robot using the closed-loop image-space MPC. Each row shows the snapshots of a plan with a different initial object pose.}
        \label{fig:real_exps}
    \end{center}
\end{figure*}

\begin{figure}[bt]
    \includegraphics[width=0.9\linewidth]{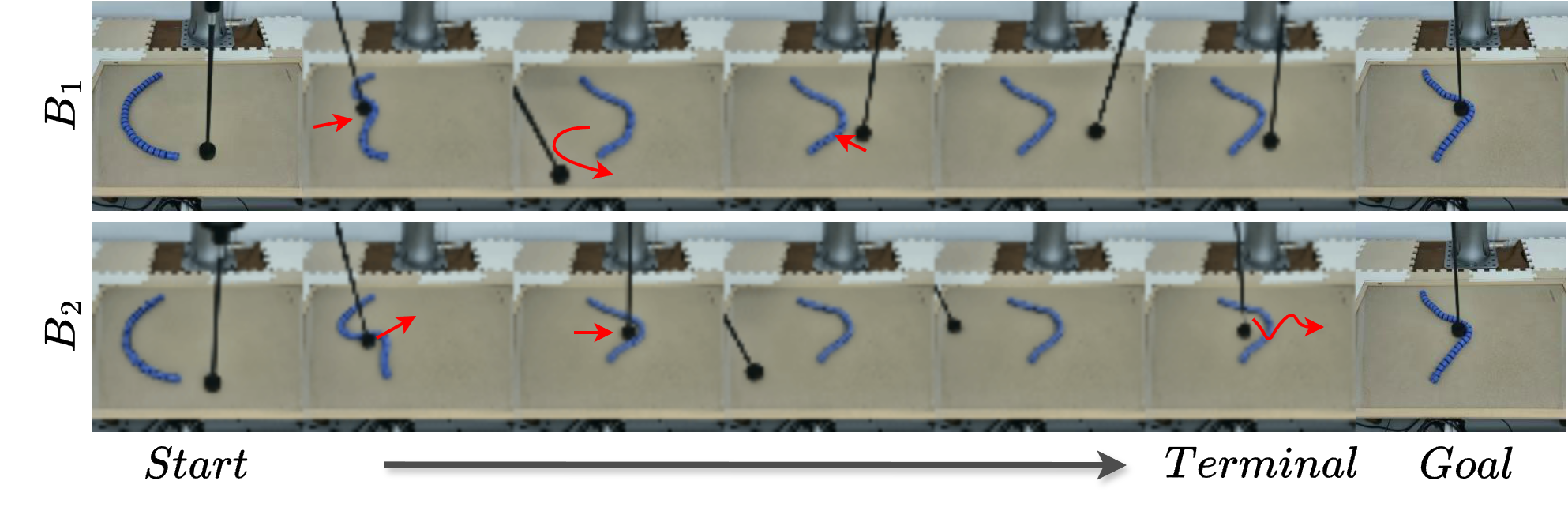}
    \caption{Two successful trajectories generated by the MCTS planner with identical start and goal states. Due to the stochasticity of the sampler and multi-modality of the diffusion world model, multiple runs can lead to multiple valid solutions ($B_1, B_2$). Red arrows are added manually to make the end effector movement easier to interpret.}
    \label{fig:rope-qualitative}
\end{figure}
\subsection{Deformable Objects Interaction}
We now use our planner to manipulate a flexible chain by pushing it until it adopts a desired form, as shown in a goal image (Fig.~\ref{fig:rollouts}, \emph{push-chain} environment). We employ the same masked DINOv2-based reward as in the previous section, masking out everything except the chain before feeding the DINOv2 encoder for distance computation. As shown in Fig.~\ref{fig:rope-qualitative}, two independent executions of MCTS may yield different strategies for solving the task. Specifically, the first run produces behavior $B_1$, in which the robot pushes the chain all the way to the right and then nudges it from the opposite side to fix the form. The second run produces behavior $B_2$, in which the chain is again pushed to the right but corrected at the end through vertical motions of the robot. The planner’s stochasticity arises both from the random action primitives sampled from a policy modeling high-entropy play behavior and from the diffusion world model, which captures the multi-modality of the observations (small perturbations during rollout can trigger a different mode).

\begin{figure}[tb]
    \centering
    \includegraphics[width=0.7\linewidth]{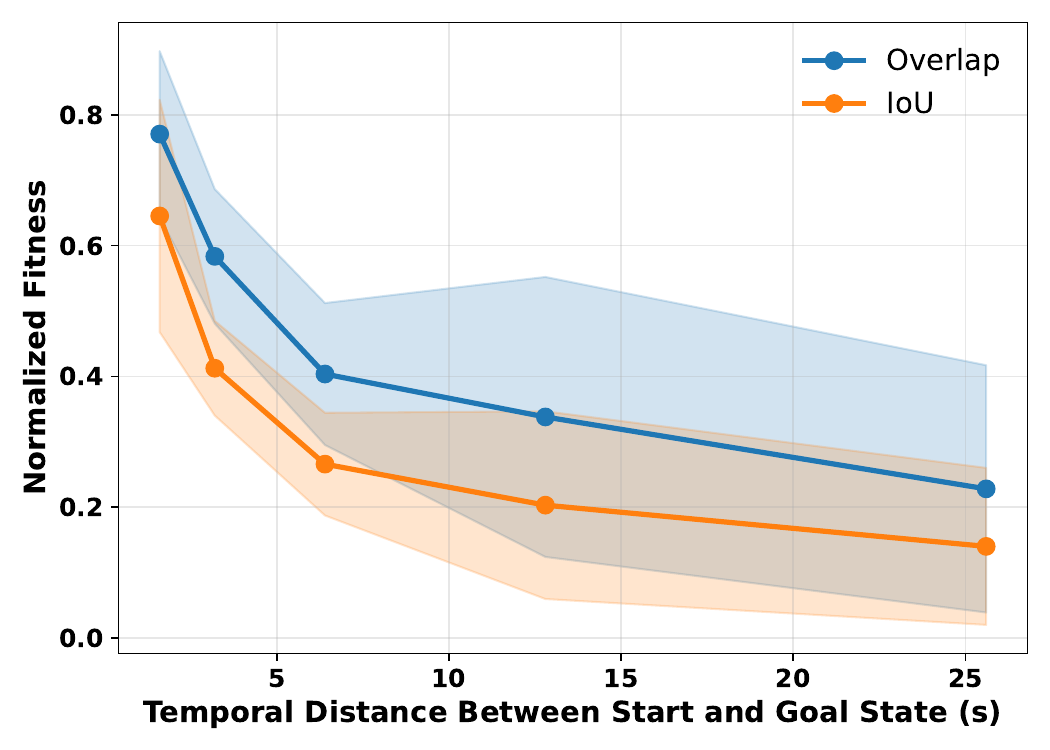}
    \caption{The similarity between the rope in the terminal states of the plan and the goal image for different problem lengths. The IoU is more sensitive to object deformation during planning.} 
    \label{fig:rope-quantitative}
\end{figure}
\subsubsection{Problem Length vs Performance}
\begin{figure}[bt]
    \centering
    \includegraphics[width=0.85\linewidth]{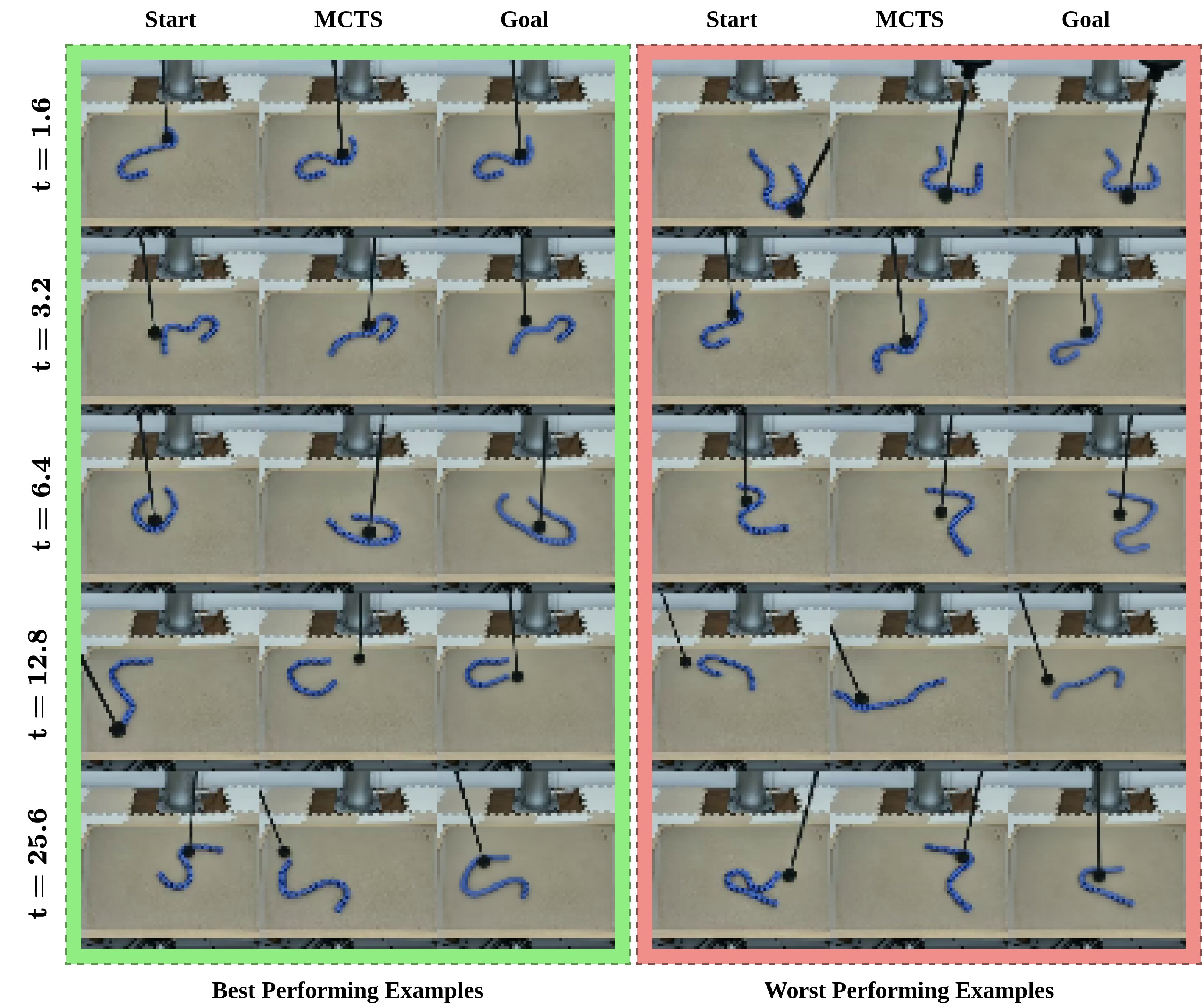}
    \caption{The world model can predict complex deformations of the chain when pushed by the robot.}
    \label{fig:rope-best-worst}
\end{figure}

In this section, videos from the play dataset are used to define random problems with various levels of difficulty. Specifically, random chunks of fixed temporal length from the dataset are considered as optimal plans, and the first and last frames are used to define the start and goal states of the planner. For the plans found for each problem, we compute two metrics: The Intersection over Union (IoU) measures how well the chain in the terminal state overlaps the chain in the goal image. This measure is sensitive to both planning performance and preservation of the object characteristics (e.g., chain length/width). Additionally, we report a coverage metric defined as the portion of the chain in the terminal state overlapping the chain in the goal state, which is less sensitive to the pattern of chain deformation in our case. Fig.~\ref{fig:rope-quantitative} shows an expected negative correlation between problem difficulty and performance. Nevertheless, as shown in Fig.~\ref{fig:rope-best-worst}, even for the worst case examples in each problem category (each row of Fig.~\ref{fig:rope-best-worst}), the terminal states of the planner are qualitatively close to the goal.   

\subsection{Closed-Loop Plan Execution with MPC}
The plans generated by MCTS can be directly tracked by an MPC controller formulated with the same world and value models (Sec. \ref{sec:method_local_planner}). With this controller, planning does not need to be repeated at every step, as MPC can tolerate small disturbances as well as modeling and execution errors. In this section, we demonstrate this integration on the \textit{push-T} task. At the beginning of each trial, the planner is triggered to find a trajectory from the current state observed by the camera to a configuration where the “T” is placed at the center of the board. Although our code is not optimized to leverage quantization or tensor cores on the GPU, plans are generated in under five minutes on average, with MCTS rollouts ranging from 80 to 300 steps. This fast convergence is due to the lightweight world model adopted in this paper and, more importantly, to the guided search strategy enabled by the action-prior policy.

After the plan is returned from MCTS, the MPC uses an unmasked version of DINOv2 distance to measure the distance to the states in the plan. Note that since the difference between observation and target is expected to be low during the MPC execution, an unmasked DINOv2 is enough and requires the least compute time among the non-geometric reward formulations presented in this paper. Each MPC iteration optimizes two knot points of a spline of length $4$ corresponding to a horizon equal to $0.8$ seconds. In our tests, we adopt a sample size of $16$ and optimize for $10$ iterations per control step (taking a couple of seconds). At the end of each MPC optimization step, we execute an action chunk of length $0.4$ seconds (2 steps) on the robot and repeat the process, see Alg.~\ref{alg:mpc}. Fig.~\ref{fig:real_exps} shows a snapshot of the real robot tracking the plan generated for three random initializations of the T-tool pose.

\section{Conclusion and Future Works}
\label{sec:conclusion}
In this paper, we propose a framework that, instead of relying on a policy to memorize dataset actions as in behavior cloning, generates robot motion through planning and control in learned visual world models. Our robotic experiments demonstrate that visual world models, trained exclusively on a few hours of goal-independent play data, enable long-horizon forward simulation and serve as the foundation for MCTS and MPC planners to synthesize new trajectories. The framework presents a promising direction for transferring knowledge across diverse tasks and supporting lifelong interaction experience. In future work, we aim to incorporate foundation joint action and world models to enable hierarchical planning and zero-shot task execution in novel settings.


\bibliographystyle{IEEEtran}
\bibliography{ref.bib}
\end{document}